# PF-DAformer: Proximal Femur Segmentation via Domain Adaptive Transformer for Dual-Center QCT

Rochak Dhakal[1], Chen Zhao[2], Zixin Shi[1], Joyce H. Keyak[3], Tadashi S. Kaneko[4], Kuan-Jui Su[5], Hui Shen[5], Hong-Wen Deng[5], Weihua Zhou[1,6]*

1. Department of Applied Computing, Michigan Technological University, Houghton, MI, 49931, USA
2. Department of Computer Science, Kennesaw State University, Marietta, GA, 30060, USA
3. Department of Radiological Sciences, Department of Mechanical and Aerospace Engineering, Department of Biomedical Engineering, and Chao Family Comprehensive Cancer Center, University of California, Irvine, Irvine, CA, 92868, USA
4. Department of Radiological Sciences, University of California, Irvine, Irvine, CA, 92868, USA
5. Tulane Center of Bioinformatics and Genomics, Tulane University School of Public Health and Tropical Medicine, New Orleans, LA, 70112, USA
6. Center of Biocomputing and Digital Health, Institute of Computing and Cybersystems, and Health Research Institute, Michigan Technological University, Houghton, MI, 49931, USA

**Abstract**

Quantitative computed tomography (QCT) plays a crucial role in assessing bone strength and fracture risk by enabling volumetric analysis of bone density distribution in the proximal femur. However, deploying automated segmentation models in practice remains difficult because deep networks trained on one dataset often fail when applied to another. This failure stems from domain shift, where scanners, reconstruction settings, and patient demographics vary across institutions, leading to unstable predictions and unreliable quantitative metrics. Overcoming this barrier is essential for multi-center osteoporosis research and for ensuring that radiomics and structural finite element analysis results remain reproducible across sites. In this work, we developed a domain-adaptive transformer segmentation framework tailored for multi-institutional QCT. Our model is trained and validated on one of the largest hip fracture related research cohorts to date, comprising 1,024 QCT images scans from Tulane University and 384 scans from Rochester, Minnesota for proximal femur segmentation. To address domain shift, we integrate two complementary strategies within a 3D TransUNet backbone: adversarial alignment via Gradient Reversal Layer (GRL), which discourages the network from encoding site-specific cues, and statistical alignment via Maximum Mean Discrepancy (MMD), which explicitly reduces distributional mismatches between institutions. This dual mechanism balances invariance and fine-grained alignment, enabling scanner-agnostic feature learning while preserving anatomical detail. Experimental results demonstrate that the combined strategy for domain adaptation using GRL and MMD yields the most consistent performance, achieving a Dice similarity coefficient of 99.53 %, and a Precision of 99.64 %, and a Hausdorff Distance of 0.77 mm in femur segmentation, all significantly improved over a non-adaptive baseline ($p < 0.01$). Beyond surface accuracy, we further show that the radiomic features extracted from adapted segmentation remain virtually identical to the ground truth (Pearson $r > 0.99$, with several > 0.9998), underscoring that fidelity is preserved across domains.

---

*Corresponding author: whzhou@mtu.edu

# 1 Introduction

Quantitative computed tomography (QCT) is a three-dimensional imaging technique that provides volumetric BMD measurements and enables separate assessment of trabecular and cortical bone, offering structural information beyond what is possible with two-dimensional (2D) dual-energy x-ray absorptiometry imaging, which is commonly used for bone density measurement [1]. QCT provides a reliable foundation for identifying critical anatomical landmarks and constructing computational models essential for hip fracture risk assessment, osteoporosis diagnosis, and monitoring of osteoporosis and other metabolic bone diseases [2]. At the center of this process is accurate segmentation of the proximal femur, which defines the region from which critical biomarkers are derived. If segmentation is unreliable, analytical results may be distorted [3], leading to incorrect risk assessment and poor clinical decisions. However, manual segmentation is not only time-consuming and inconsistent but also irreproducible across operators, making automation indispensable for both clinical deployment and large-scale research [4,5].

Automatic segmentation of QCT volumes presents unique challenges due to subtle tissue contrast, high class imbalance, and heterogeneous acquisition protocols. More importantly, QCT datasets are typically collected across different institutions, scanners, populations, and reconstruction settings, leading to substantial intra-modality variability. This domain shift can degrade the performance of deep learning models [6] trained on a single source dataset because deep learning models work best when the training and testing data come from the same distribution. Their performance drops when this consistency is lost, highlighting the need for strategies that ensure robustness across multi-center QCT data.

More recently, hybrid models that integrate transformers have shown promise for volumetric tasks. Architectures such as TransUNet3D [7], SWIN Transformer [8] and SegFormer3D [9] uses vision transformers (ViTs) for modeling long-range dependencies, achieving state-of-the-art results in segmentation. These models demonstrate that combining local convolutional representations with global self-attention can capture both fine anatomical detail and broader contextual information, which is particularly valuable for volumetric QCT segmentation. Building on this foundation, transformer-based approaches offer a powerful backbone for domain-adaptive QCT analysis.

To address the critical challenge of inter-domain variability in QCT, our model integrates Gradient Reversal Layer (GRL) with Maximum Mean Discrepancy (MMD) GRL enforces scanner-invariant feature learning by adversarially discouraging site-specific cues, providing coarse alignment across domains. This forces the network to learn features that cannot distinguish between source and target domains, thereby promoting domain invariance [10]. However, this adversarial setup only ensures that the classifier is confused, and it does not measure or minimize the actual statistical distance between domains. GRL over looks subtle but clinically important mismatches, such as differences in intensity scaling or bone density distributions, because these do not strongly influence the domain classifier's decision boundary. MMD directly complements this by measuring and minimizing the statistical distance between source and target distributions in a high-dimensional kernel space, thereby capturing higher-order discrepancies [11] that GRL cannot. MMD alone, however, lacks adversarial pressure and may fail to remove scanner artifacts or confounding site-specific patterns. By combining them, our proposed Proximal Femur Domain Adaptation with Transformer (PF-DAformer) unites GRL's broad invariance with MMD's fine-grained statistical alignment, achieving robust cross-site generalization that neither method can deliver independently.

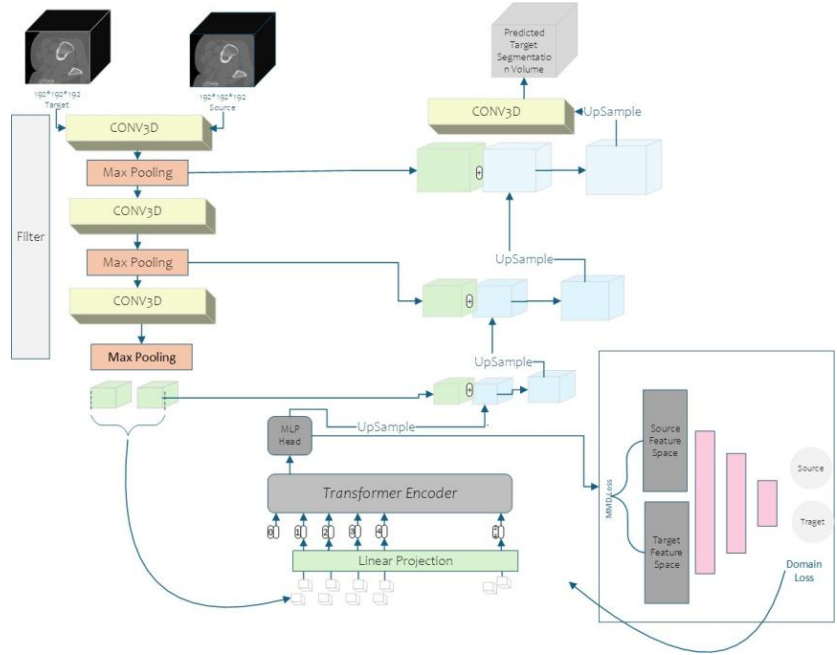

*Figure 1: The proposed PF-DAformer segmentation model integrates a CNN encoder for local feature extraction, a Vision Transformer for global context modeling, and a decoder with skip connections for precise segmentation. A Gradient Reversal Layer (GRL) enables adversarial domain adaptation by aligning source and target feature spaces, ensuring accurate and domain-invariant QCT segmentation.*

Compared to alternatives such as Correlation Alignment (CORAL) [12], which is limited to aligning first-order and second-order statistics, or style transfer methods that can destabilize training through image-level adaptation, our feature-level hybrid strategy is both more stable and better suited for multi-institutional QCT, where reliability and reproducibility are paramount. In addition, we develop a strong transformer baseline tailored for 3D QCT segmentation. By carefully evaluating and fine-tuning state-of-the-art encoder decoder architectures, we establish a high-fidelity baseline capable of preserving detailed trabecular and cortical structures. This serves as a rigorous reference to isolate the impact of our domain adaptation module.

**Our main contributions are summarized as follows:**

1. **Hybrid Domain Adaptation:** We combine GRL and MMD for complementary adversarial and statistical alignment, improving cross-site generalization in multi-institutional QCT.
2. **Baseline Development:** We fine-tune transformer encoder-decoder models to create a strong 3D QCT segmentation baseline for fair and rigorous evaluation.

## 2   Related Works

Deep learning has substantially advanced medical image segmentation. A study from Zhao et al. introduced ST-V-Net [13], which integrates shape prior information into convolutional architectures for proximal femur segmentation on a single-center QCT dataset, demonstrating improved boundary accuracy and anatomical plausibility. While effective within a controlled dataset, this work did not address inter-institutional variability.

While network architecture continues to evolve, the problem of domain shift remains central to real-world deployment. Domain adaptation (DA) techniques aim to bridge the distribution gap between source and target datasets by learning domain-invariant yet task-discriminative representations. Much of the early work in medical imaging focused on cross-modal adaptation, particularly between MRI and CT. For example, Chen *et al.* introduced SIFA [14],introduced Synergistic Image and Feature Adaptation (SIFA), a method that addresses the domain shift from two sides. First, it transforms source images so that they look like the target modality, reducing the obvious style differences between scans. At the same time, it uses adversarial learning in the feature space so that the model learns similar internal representations for both domains. By combining these two steps, SIFA builds a bridge between modalities and achieves strong segmentation performance without the need for labels in the target domain.

Ouyang *et al.* [15] demonstrated that adversarial training enables 3D cardiac segmentation between MRI and CT with limited supervision. Xian *et al.* employed DADASeg-Net with dual adversarial attention, using spatial and class attention maps, to achieve more effective cross-modality medical image segmentation. These studies highlight the value of adversarial strategies for overcoming domain gaps. While successful in those contexts, they do not address the more subtle but equally damaging problem of intramodal variability.

Intra-modality domain adaptation has only recently gained attention. Brion *et al.* [16]showed that domain adversarial networks and intensity-based data enhancements improved each robustness when adapting between CT and cone beam CT. Similarly, Chen *et al.* [17] addressed the problem of low-dose versus normal-dose CT by introducing an unsupervised adaptation framework that integrates a Fourier-based UNet (F-UNet) with a Weighted Segmentation Re- construction (WSR) module. This design improves frequency domain alignment and segmentation accuracy, highlighting that even subtle intramodality changes, such as dose variations, can significantly degrade performance if not addressed. In the context of QCT, Zhang *et al.* introduced DeepmdQCT [18], a framework that leverages domain-invariant feature learning and attention mechanisms for osteoporosis diagnosis and bone density estimation, highlighting the clinical feasibility of domain adaptation in QCT analysis.

Different families of DA strategies have been explored in medical imaging. Adversarial methods, such as gradient reversal layers (GRLs), introduce a domain classifier whose gradients are inverted during training, forcing the encoder to learn features that are indistinguishable across domains [10]. Discrepancy-based approaches, including maximum mean discrepancy (MMD) [11] and correlation alignment (CORAL), explicitly minimize the statistical distance between source and target feature distributions. Despite these advances, empirical evidence consistently indicates that combining complementary adaptation strategies is more effective than relying on any single method. For instance, adversarial learning promotes invariance but may overlook fine distributional mismatches, while MMD captures higher-order discrepancies but lacks explicit adversarial regularization.

Thus, our study introduces a hybrid CNN based Transformer segmentation model with integrated adversarial (GRL) and discrepancy-based (MMD) adaptation. By explicitly leveraging annotations across two QCT datasets and aligning their feature distributions, we aim to achieve scanner-invariant and anatomically precise segmentation. Our contributions demonstrate that domain adaptation not only improves overlap-based accuracy but also enhances boundary robustness, paving the way for reliable and generalizable QCT analysis in the multi-institutional settings.

# 3 Data Acquisition and Preprocessing

The dataset enrolled comprises 1,024 QCT scans collected at Tulane University and 384 QCT scans from a study conducted in Rochester, Minnesota, USA. Each volume contains between 37 and 95 slices. Ground truth annotations are provided exclusively for the left proximal femur, which appears on the right side of each axial slice from the patient's perspective.

## 3.1 Comparative Context of QCT Datasets

The Tulane dataset is closely aligned with imaging protocols from the Louisiana Osteoporosis Study (LOS), a large cohort study designed to investigate genetic and environmental risk factors for osteoporosis and musculoskeletal disorders [19] [20]. In this study, QCT scans were obtained on the GE Discovery CT750 HD system at Tulane University Department of Radiology. The LOS subset included 1024 males aged 20-50 years (African American and Caucasian). The imaging protocol used 2.5 mm slice thickness with the pixel sizes between 0.695-0.986 mm, and a $512 \times 512$ matrix, providing a rich resource for multi-modal investigations of bone fragility.

Another QCT dataset from a Rochester, Minnesota cohort, which contains QCT scans from 397 individuals (216 females, 181 males) aged 27-90 years, predominantly White (>95%)[13] [21]. These QCT scans were acquired using a Siemens Sensation 64 system at 120 kVp with 2 mm slice thickness, later reconstructed to 3 mm slices using Fourier interpolation to ensure consistency with finite element structural analysis (FE) protocols. QCT images were reconstructed to 512×512 resolutions, with pixel sizes ranging from 0.742-0.977 mm and processed with a B30s convolutional kernel. This dataset has primarily been used to study structural and mechanical properties of the proximal femur in middle-aged and elderly populations.

Together, these datasets illustrate the diversity of QCT imaging protocols and study populations.

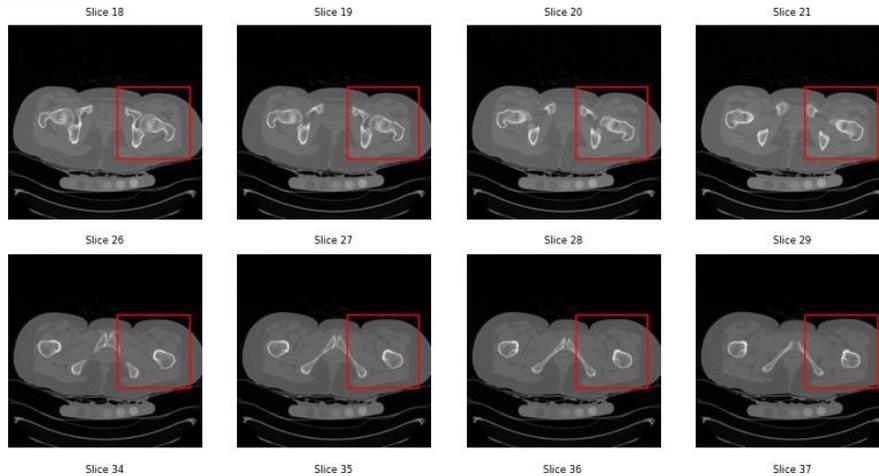

*Figure 2: Illustration of cropping operation applied to each axial slice. The red box highlights the extracted region corresponding to the left proximal femur.*

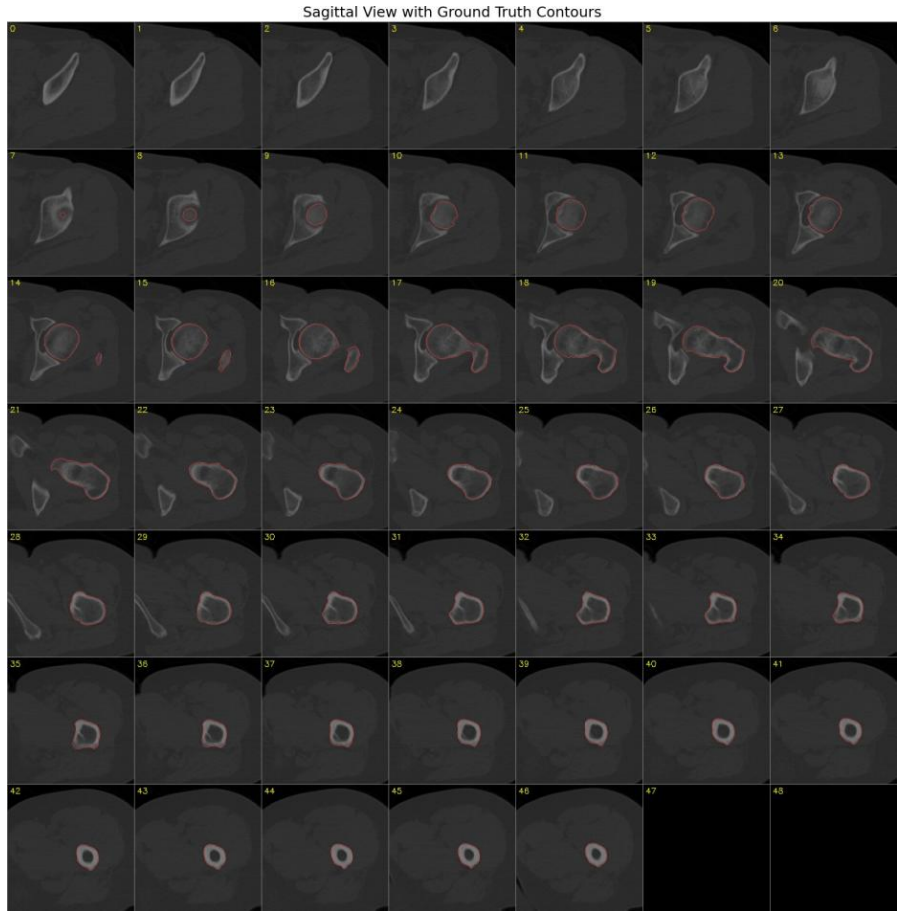

*Figure 3: Sagittal view of the volume with ground truth contours overlaid. The highlighted regions represent the manually annotated segmentation boundaries for the left proximal femur.*

## 3.2 Data Preprocessing of QCT Datasets

First, we cropped a rectangular region from the middle-right part of each slice to isolate the left proximal femur, as shown in Figure 4. This crop focuses computational resources on the annotated region, reducing memory requirements and excluding irrelevant anatomical structures to improve training efficiency.

Second, to handle variable slice counts and ensure compatibility with our 3D TransUNet-based model, we standardize both the cropped volumes and labels to a cubic size of 192×192 ×.192 This size aligns with the patch-based processing of the Vision Transformer (ViT) encoder, ensuring consistent input dimensions. The standardization process involves symmetrically padding with zeros for volumes and labels with fewer than 192 slices or centrally cropping those exceeding 192 slices to retain relevant anatomical content. This uniform size facilitates efficient batch processing and compatibility with the model's fixed-size layers, despite a minor computational overhead from padding, enabling robust performance across diverse QCT datasets.

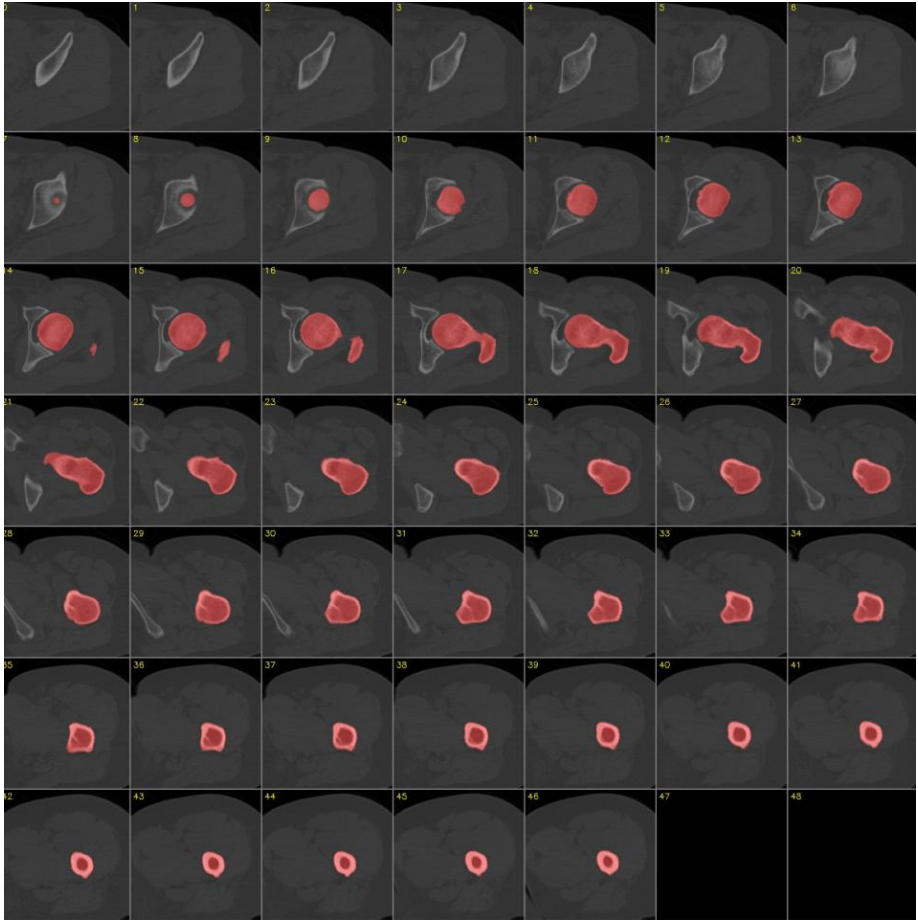

*Figure 4: Tiled sagittal slices of the proximal femur with ground truth segmentation overlays. The red filled regions represent the annotated anatomical structures, overlaid on grayscale intensity slices. Slice indices and grid lines are included for spatial reference.*

The preprocessed volumes and labels are stored as tensors with a shape of 1×192×192×192, accommodating the single-channel QCT data. This pipeline ensures efficient and consistent training for the DA model.

## 4 Methodology

### 4.1 Overview of 3D Model Architecture

Our model is inspired by the TransUNet3D's encoder framework [7] for the segmentation of volumetric QCT data. The model integrates a CNN-based encoder for local feature extraction, a 3D ViT for global contextual modeling, and a decoder with skip connections for precise segmentation. To mitigate domain discrepancies, we incorporate adversarial domain adaptation using a GRL, enabling domain-invariant feature learning. The training combines supervised segmentation losses with adversarial domain losses, enhancing both segmentation accuracy and cross-domain generalizability for 3D QCT volumes.

In this paper, we aim to improve the segmentation performance on the Tu- lane (source) dataset by learning domain-invariant features through adversarial alignment with the Rochester QCT-Femur (target) dataset. Although prior work has demonstrated strong performance on

Rochester QCT-Femur[13], our objective is to exploit Rochester QCT-Femur images to regularize the feature space such that Tulane and Rochester QCT-Femur distributions are aligned, thereby enhancing generalization on Tulane's dataset.

The architecture comprises four main components: a 3D CNN encoder, a 3D ViT encoder, a domain adaptation block and a 3D decoder. Below, we describe each component and the domain adaptation strategy in detail.

### 4.1.1 3D CNN Encoder

Let the input QCT volume be represented as $X \in R^{1 \times D \times H \times W}$, where $D$, $H$, and $W$ are the depth, height, and width ($192 \times 192 \times 192$). The encoder applies three sequential 3D convolutional blocks to extract local anatomical features while progressively reducing spatial dimensions. Each block consists of two

$3 \times 3 \times 3$ convolutional layers, each followed by batch normalization (BN) and a ReLU activation. A $2 \times 2 \times 2$ max pooling layer is applied between blocks, halving the spatial dimensions at each stage. This produces feature volumes of resolutions $D/2 \times H/2 \times W/2$, $D/4 \times H/4 \times W/4$, and $D/8 \times H/8 \times W/8$, with channel depths of 32, 64, and 128, respectively.

### 4.1.2 3D Vision Transformer Encoder

The compressed CNN feature map ($X_{cnn} \in R^{128 \times 24 \times 24 \times 24}$) is partitioned into non-overlapping $8 \times 8 \times 8$ patches, yielding.

$$N = \left(\frac{24}{8}\right)^3 = 27 \text{ patches per volume.}$$

Each patch is linearly projected into a latent space of dimension $d = 512$ using a 3D convolutional layer with kernel and stride equal to the patch size. Learnable

positional embeddings are then added to preserve spatial context:

$$x_{patch} = \text{Conv3D}(x_{cnn}) + E_{pos} \quad (1)$$

where $E_{pos} \in R^{1 \times N \times d}$ are the positional embeddings.

The 3D ViT consists of six transformer blocks, each comprising layer. normalization, multi-head self-attention, and a multilayer perceptron (MLP) with hidden size 2048. Residual connections are applied in both attention and MLP layers. For each head, the query (Q), key (K), and value (V) matrices are computed as:

$$Q, K, V \in R^{N \times d_h}, d_h = \frac{d}{h} = \frac{512}{8} = 64,$$

where $N = 27$ is the number of patches and $h = 8$ is the number of heads. The self-attention operation is then

$$\text{Attention}(Q, K, V) = \text{Softmax}\left(\frac{QK^T}{\sqrt{d_h}}\right) \quad (2)$$

The outputs of all heads are concatenated and reshaped back into a 3D feature volume ($512 \times 24 \times 24 \times 24$) for decoding.

### 4.1.3 Adversarial Domain Adaptation

A major source of performance degradation in multi-institutional QCT arises from domain shift, where models trained on one dataset fail to generalize to another due to differences in scanner type, reconstruction kernel, or patient population. To mitigate this, we employ adversarial domain adaptation based on a Gradient Reversal Layer (GRL). The GRL enables the encoder to learn domain-invariant features by explicitly discouraging representations that contain site-specific cues.

The domain classifier is trained with the standard cross-entropy loss:

$$L_{adv} = -\frac{1}{B}\sum_{i=1}^{B}\sum_{c\in\{src,tgt\}} d_{i,c} \log D(f_i)_c,$$

where $d_{i,c}$ is the one-hot domain label for sample $i$. Here, $f_i$ denotes the feature representation extracted by the encoder for the $i^{th}$ input sample, i.e., $f_i = E(x_i)$, where $E(\cdot)$ is the encoder network and $x_i$ is the corresponding input image from either the source or target domain. The function $D(\cdot)$ represents the domain classifier, which takes fi as input and predicts the probability distribution over domain labels (source vs. target) using SoftMax activation. Specifically, D(fi)c corresponds to the predicted probability that the feature $f_i$ originates from domain $c \in \{src, tgt\}$.

Without gradient reversal, both the encoder and the domain classifier would minimize, encouraging the encoder to produce features that make domains more distinguishable. The GRL changes this dynamic: during backpropagation the gradient received by the encoder is multiplied by $-\lambda$, so the encoder update becomes $\Delta\theta_{enc}$, as shown in Eq. 4.

$$\Delta\theta_{enc} \propto -\frac{\partial L_{adv}}{\partial f} \cdot \frac{\partial f}{\partial \theta_{enc}} \cdot (-\lambda) = +\lambda \frac{\partial L_{adv}}{\partial f} \cdot \frac{\partial f}{\partial \theta_{enc}}, \qquad (4)$$

which is equivalent to maximizing $L_{dom}$ with respect to the encoder parameters $\theta_{enc}$. In other words, the domain classifier attempts to distinguish source from target, while the encoder simultaneously learns to generate representations that confuse it.

This adversarial game drives the encoder towards features that are domain-invariant yet still optimized for segmentation. In PF-DAformer, the GRL is applied after the ViT encoder but before the decoder. This placement is intentional: the ViT-encoded features capture both local bone morphology and global anatomical context, making them an ideal point to enforce domain invariance. By reversing gradients from the domain classifier at this stage, the encoder is trained to suppress scanner-specific and site-specific variations, while the decoder simultaneously optimizes segmentation performance on the source labels. This adversarial interplay ensures that the final segmentation head operates on features that are both anatomically informative and robust across institutions. Formally let $f = \mathbf{R}^{B \times d \times H' \times W' \times D'}$ denote the feature volume extracted by the ViT encoder ($d = 512$, $H' = W' = D' = 24$). During forward propagation,

the GRL acts as the identity mapping:

$$\text{GRL}(f) = f, \qquad (5)$$

but during backpropagation, it multiplies the gradient by a negative scalar $\lambda$:

$$\frac{\partial \text{GRL}(f)}{\partial f} = -\lambda I, \qquad (6)$$

where $\lambda > 0$ controls the strength of the adversarial signal and $I$ is the identity matrix. The features from the ViT encoder are routed through the GRL before being passed to a lightweight domain classifier. Although the GRL does not modify the features in the forward path, it ensures that the gradients arriving from the domain classifier are reversed when propagated back to the encoder. The domain classifier itself attempts to predict whether a given

feature volume originates from the source or target dataset:

$$D(f) = \sigma(W_d f + b_d), \tag{7}$$

where $W_d$ and $b_d$ denote the classifier weights and bias, and $\sigma( )$ is the sigmoid activation. To reduce the spatial dimension, we first apply global average pooling:

$$f_{gap} = GAP(f) \in R^{B \times 512} \tag{8}$$

where $B$ is the batch size. This pooled feature is then passed through fully connected layers with dimensions $512 \rightarrow 128 \rightarrow 64 \rightarrow 2$, each followed by ReLU, layer normalization, and dropout ($p = 0.2$). The final layer outputs logits for the two domain classes (source vs. target). Adversarial training thus enforces the encoder to suppress site-specific cues while preserving features critical for segmentation.

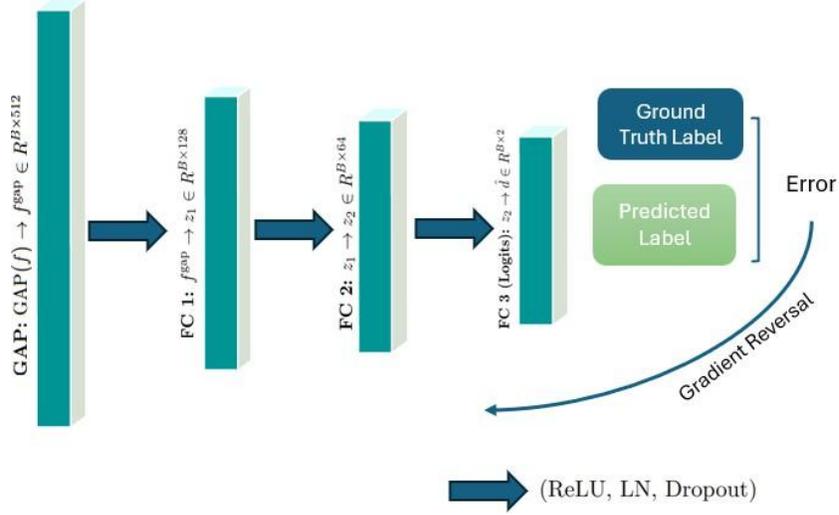

Figure 5: GAP and Fully Connected (FC) Domain Classifier Head. The figure illustrates the lightweight domain classification branch used in adversarial domain adaptation. The feature volume is first aggregated using global average pooling (GAP), producing a 512-dimensional vector. This is followed by three fully connected (FC) layers that progressively reduce the feature dimensionality ($512 \rightarrow 128 \rightarrow 64 \rightarrow 2$) The outputs correspond to the domain logits for source vs. target classification, which are used in the adversarial loss to encourage domain-invariant feature learning. Arrow labels ($f^{gap}$, $z_1$, $z_2$) represent inter-mediate vector representations between layers. Box labels indicate the layer type and input/output dimensions.

#### 4.1.4 3D Decoder

The decoder reconstructs the segmentation output from the ViT-encoded features, progressively upsampling to the original resolution using trilinear interpolation. Skip connections concatenate encoder features from corresponding resolutions to recover spatial details. Each decoder block applies two 3×3×3 convolutional layers with BN and ReLU, defined as:

$$x_d = \text{Conv3D}\big(\text{Concat}(\text{Up}(x_e), x_s)\big), \tag{9}$$

where $x_e$ is the encoded feature, $x_s$ is the skip-connected feature, and Up denotes trilinear upsampling. The final segmentation prediction $\hat{y}$ is obtained through a 1 × 1 × 1 convolutional layer followed by a SoftMax activation function, producing a voxel-wise probability map over all segmentation classes:

$$\hat{y} = \text{Softmax}\big(\text{Conv3D}_{1\times1\times1}(x_d)\big). \tag{10}$$

It is important to note that the decoder is not directly linked to the domain classifier. Both branches operate in parallel from the same ViT-encoded feature space: one branch is directed through the GRL and domain classifier to enforce domain invariance, while the other branch is decoded into a segmentation mask. The gradient reversal mechanism ensures that updates from the domain classifier indirectly influence the encoder, shaping the features shared with the decoder. In this way, the decoder benefits from more robust, domain-invariant features without requiring any explicit interaction with the domain classification head.

## 4.2 Loss Functions

The integration of composite loss function has proven transformative [22] [23] for PF-DAformer, significantly enhancing its performance and robustness across diverse datasets. Unlike conventional approaches relying solely on Dice or cross-entropy (CE) losses, our approach integrates baseline Loss (Dice + CE), Focal Loss, Domain Loss, and Maximum Mean Discrepancy (MMD) Loss into a cohesive framework. This strategic combination has emerged as a game changer, dramatically improving segmentation accuracy and enabling superior domain generalization.

### 4.2.1 Segmentation Loss

Our model employs a composite segmentation loss that integrates three complementary components: Dice loss, binary CE loss, and focal loss. This combination is designed to ensure accurate delineation of anatomical structures, stable pixel-wise supervision, and enhanced sensitivity to hard to classify pixels, particularly in the presence of severe foreground and background imbalance. Some studies reported improved segmentation performance when combining Dice with CE as well as Dice with Focal loss. In contrast, we combined all three losses given their proven effective individually.

The base loss consists of Dice loss and CE loss, weighted equally. Dice loss directly measures the overlap between predicted and ground-truth regions, making it well-suited for handling class imbalance and improving recall for small structures. CE loss, on the other hand, provides pixel-level supervision that stabilizes optimization and encourages accurate classification on a per-pixel basis. Let $(p_i \in [0,1])$ denote the predicted probability for pixel $(i), and (y_i \in \{0,1\})$ the corresponding ground-truth label, for a total of $N$ pixels. Dice loss is defined as:

$$L_{\text{Dice}} = 1 - \frac{2\sum_{i=1}^{N} p_i y_i}{\sum_{i=1}^{N} p_i + \sum_{i=1}^{N} y_i}, \tag{11}$$

The CE loss is expressed as:

$$L_{CE} = -\frac{1}{N}\sum_{i=1}^{N}[y_i \log(p_i) + (1 - y_i) \log(1 - p_i)] \tag{12}$$

The base loss combines these two terms as:

$$L_{\text{base}} = L_{\text{Dice}} + L_{\text{CE}}. \tag{13}$$

This combination enables the model to learn both global structure (via Dice) and local pixel-wise accuracy (via CE), which is particularly important for extracting the anatomical regions with varying sizes and contrasts.

In dense prediction settings such as image segmentation, the number of background (negative) pixels typically far exceeds that of foreground (positive) pixels. Most background pixels are easy to classify, and their dominance in the training set can cause the loss to be driven by these easy negatives, reducing the model's ability to learn from harder, positive pixels.

To mitigate this, we incorporate the Focal Loss introduced by Lin *et al.* [24],which down-weights well-classified pixels and emphasizes hard examples. Starting from the binary cross-entropy formulation, a weighting factor $\alpha$ [0, 1] is introduced to balance positive and negative classes:

$$\text{CE}_\alpha(p_i, y_i) = -\alpha y_i \log(p_i) - (1-\alpha)(1-y_i)\log(1-p_i). \qquad (14)$$

Next, a modulating factor $(1 - p_{t,i})^\gamma$ is applied, where controls the extent to which easy examples are downweighed. Here,

#cooment

The focal loss for each pixel is given by:

$$\text{FL}(p_{t,i}) = -\alpha_i(1 - p_{t,i})^\gamma \log(p_{t,i}), \qquad (15)$$

and the total focal loss is averaged across all pixels

$$L_{\text{Focal}} = \frac{1}{N}\sum_{i=1}^{N} \text{FL}(p_{t,i}) \qquad (16)$$

For well-classified pixels ($p_{t,i} \approx 1$), the modulating factor becomes small, reducing their contribution to the total loss. For hard pixels ($p_{t,i}$ low), this factor remains large, thereby focusing the loss on more challenging examples.

When ($\gamma = 0$), the focal loss reduces to the standard $\alpha$-balanced CE loss.

The final segmentation loss used to train our model combines the base loss and focal loss using a weighting parameter $\alpha \in [0.3, 0.5]$, which balances training stability with the emphasis on hard examples:

$$L_{\text{seg}} = (1-\alpha)(L_{\text{Dice}} + L_{\text{CE}}) + \alpha\, L_{\text{Focal}}. \qquad (17)$$

### 4.2.2 Domain Classifier Loss

To encourage the encoder to learn domain-invariant representations, we train a domain classifier on top of the ViT feature vectors to predict the dataset of origin (source vs. target), while the encoder simultaneously learns to make this prediction difficult through the Gradient Reversal Layer (GRL).

Let($d_i \in \{0,1\}$) denote the domain label for the $i^{\text{th}}$ input volume in a mini batch, where $d_i = 0$ corresponds to the source domain and $d_i = 1$ to the target domain. Let ($\hat{d}_i \in [0,1]$) be the predicted probability that the volume belongs to the target domain produced by the domain classifier. The domain classifier is trained using the standard binary cross-entropy loss:

$$L_{\text{adv}} = -\frac{1}{N_b}\sum_{i=1}^{N_b}[d_i \log(\hat{d}_i) + (1-d_i)\log(1-\hat{d}_i)] \qquad (18)$$

where $N_b$ is the number of volumes in the mini batch.

During backpropagation, the GRL inverts the gradient of $L_{\text{adv}}$ before it reaches the encoder, forcing the encoder to produce features that confuse the domain classifier. It is important to note that since the domain classification involves only two categories (source vs. target), the general SoftMax cross-entropy loss introduced in Eq. (3) simplifies to a binary cross-entropy formulation.

### 4.2.3 Discrepancy Alignment via Maximum Mean Discrepancy (MMD) Loss

To explicitly align feature distributions across institutions, we employ the Maximum Mean Discrepancy (MMD) loss. Let

$$F_{\text{ViT}} \in R^{B \times 512 \times 24 \times 24 \times 24}$$

denote the feature volume extracted by the ViT encoder for a batch of $B$ volumes. We apply global average pooling (GAP) to obtain a single 512-dimensional feature vector per scan:

$$f^{(i)} = \text{GAP}(F_{\text{ViT}}^{(i)}) \in R^{512}$$

For each batch, we divide the pooled features into source and target subsets:

$$\mathcal{F}_s = \{f_s^{(1)}, \dots, f_s^{(n_s)}\}, \mathcal{F}_t = \{f_t^{(1)}, \dots, f_t^{(n_t)}\},$$

where $n_s$ and $n_t$ are the number of source and target samples, respectively.

MMD measures the distance between two distributions in a reproducing kernel Hilbert space (RKHS). Intuitively, if the source and target feature distributions match, the MMD value becomes small. Using a characteristic kernel

$k(\cdot, \cdot)$ (here, a mixture of Gaussian kernels), the squared MMD is defined as

$$\text{MMD}^2(P_s, P_t) = E_{f_s, f_s'}[k(f_s, f_s')] + E_{f_t, f_t'}[k(f_t, f_t')] - 2 E_{f_s, f_t}[k(f_s, f_t)].$$

In practice, we estimate this in each mini batch using the unbiased estimator:

$$\widehat{\text{MMD}^2} = \frac{1}{n_s(n_s-1)} \sum_{i \neq i'} k\left(f_s^{(i)}, f_s^{(i')}\right)$$

$$+ \frac{1}{n_t(n_t-1)} \sum_{j \neq j'} k\left(f_t^{(j)}, f_t^{(j')}\right)$$

$$- \frac{2}{n_s n_t} \sum_{i=1}^{n_s} \sum_{j=1}^{n_t} k\left(f_s^{(i)}, f_t^{(j)}\right)$$

(19)

The first two terms measure within-domain similarity (source-source and target-target), while the last term measures cross-domain similarity. If the source and target feature distributions differ, the MMD value increases; minimizing it encourages the encoder to learn domain-invariant features.

For the kernel, we use a Gaussian RBF kernels with different bandwidths ($M = 5$) to capture discrepancies between source and target features across multiple scales:

$$k(f, f') = \sum_{m=1}^{M} \exp\left(-\frac{|f - f'|^2}{2\sigma_m^2}\right),$$

where the bandwidths $\{\sigma_m\}$ are chosen relative to the median pairwise distance in the current batch:

$$\{\sigma_m\} \in \left\{\frac{\tilde{\sigma}}{4}, \frac{\tilde{\sigma}}{\sqrt{2}}, \tilde{\sigma}, \sqrt{2}\tilde{\sigma}, 2\tilde{\sigma}\right\}$$

Using multiple $\sigma_m$ values allows the kernel to capture both fine and coarse distributional differences without manually tuning a single bandwidth. The gradient of a single Gaussian

kernel with respect to $f_s$ is:

$$\frac{\partial k_\sigma(f_s, f_t)}{\partial f_s} = \frac{1}{\sigma^2} k_\sigma(f_s, f_t)(f_t - f_s),$$

which shows that features are pulled toward their cross-domain counterparts with a strength determined by the kernel similarity and bandwidth.

Finally, the MMD loss is incorporated into the overall training objective:

$$L = L_{\text{seg}} + \alpha\, L_{\text{adv}} + \beta\, \widehat{\text{MMD}}^2,$$

where $L_{\text{seg}}$ is the supervised segmentation loss, $L_{\text{adv}}$ is the adversarial domain loss from GRL, and MMD aligns the source and target feature distributions. Together, these terms encourage the network to produce features that are simultaneously discriminative for segmentation and invariant to site-specific differences.

## 5 Results

An ablation study was conducted to evaluate different segmentation loss weightings and domain adaptation strategies. The comparison of results using different hyperparameter across all methods was statistically analyzed using ANOVA, and the results indicated that the configuration with **0.6 DiceCE loss weight and 0.4 Focal loss weight** achieved the best overall performance. This configuration was therefore selected for the main analysis.

Table 1: Comparison of segmentation performance across different methods. Best values per metric are highlighted in bold.

| Methods | Dice (%) | Precision (%) | Recall (%) | HD (↓) | HD95 (↓) | ASD (↓) |
|---|---|---|---|---|---|---|
| **Base Models (No DA)** | | | | | | |
| Swin Transformer | 95.6136 | 95.3343 | 95.9035 | 18.5492 | 1.5734 | 0.7735 |
| SegFormer3D | 98.3301 | 98.2663 | 98.4120 | 5.2822 | 1.6572 | 0.3714 |
| TransUNet | **99.5061** | **99.5682** | **99.4452** | 3.9686 | 0.8588 | 0.0902 |
| **Domain Adaptation on TransUNet** | | | | | | |
| DA (MMD only) | 99.5068 | 99.6036 | 99.4106 | 5.1711 | 0.9206 | 0.0668 |
| DA (GRL only) | 99.5250 | 99.6335 | 99.4172 | **2.3355** | 0.8373 | 0.0627 |
| DA (GRL + MMD) | **99.5307** | **99.6369** | **99.4252** | 3.1313 | **0.7706** | **0.0622** |

The results show that while all methods achieved similarly high Dice scores above 99.5%, domain adaptation strategies generally improved precision and boundary accuracy compared to the non-adaptive baseline. The combined MMD+GRL approach with weighting yielded the most balanced performance, achieving the highest Dice and precision while simultaneously reducing HD, HD95, and ASD. Gradient reversal also improved Hausdorff distance compared to the baseline but did not consistently lower HD95 or surface distance. MMD alone provided modest gains in Dice and precision but underperformed in boundary metrics, with higher HD and HD95 values.

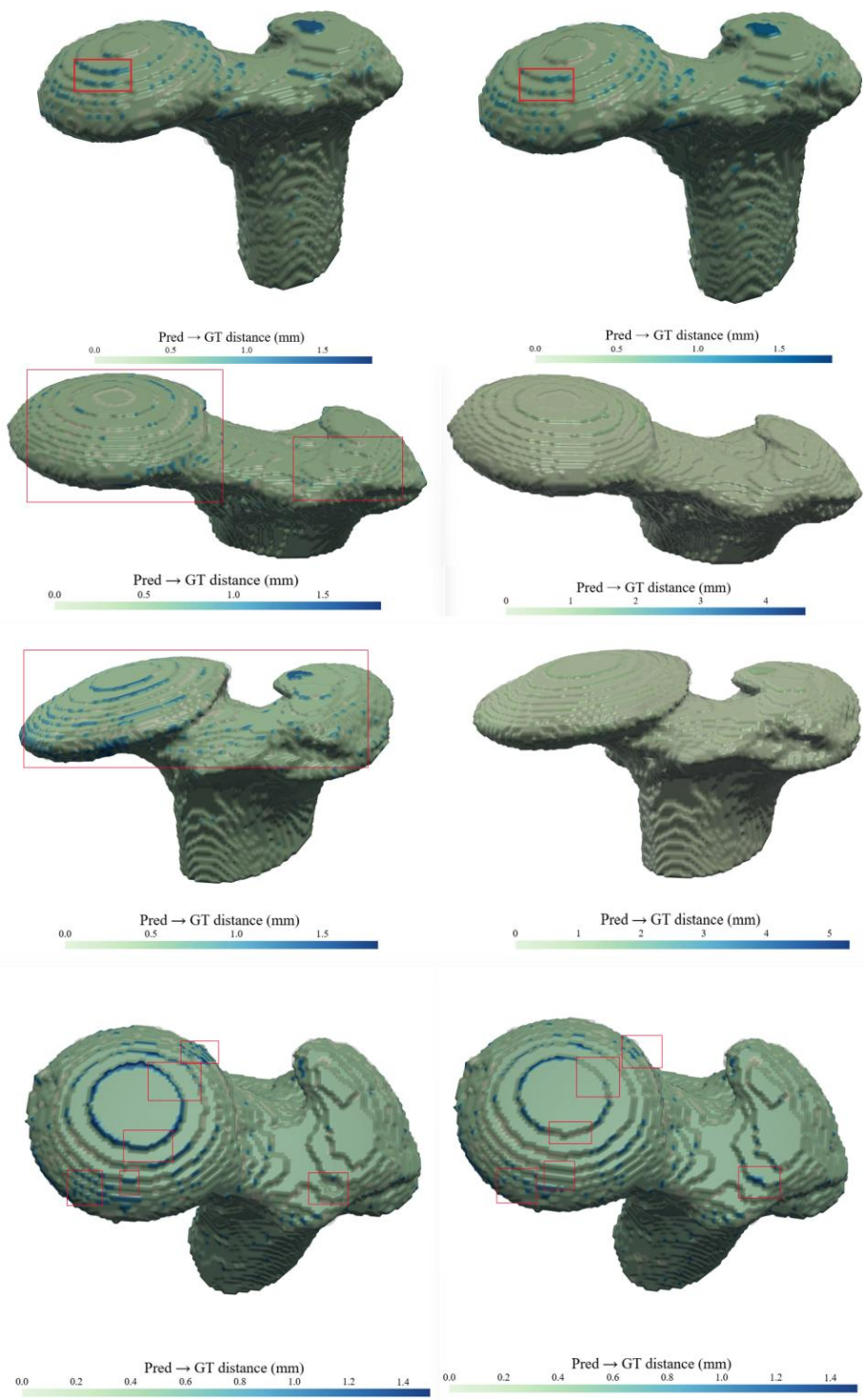

*Figure 6: Comparison of predicted surface distance to ground truth before (left) and after (right) applying domain adaptation (DA). The color map encodes the surface distance from the predicted segmentation to the GT in millimeters, with lower values (green) indicating closer alignment and higher values (blue) reflecting larger deviations. The left panel*

*(before DA) shows larger areas of discrepancy, particularly in the highlighted red box region. After incorporating DA (right panel), the deviations are reduced, as indicated by the overall shift toward lower distance values. The red box highlights a region where improvement is most apparent, demonstrating how DA reduces local surface mismatches and improves boundary alignment.*

To further quantify the effectiveness of the proposed domain adaptation (DA) strategy against the non-adaptive baseline, we performed paired *t*-tests across 205 common cases and multiple evaluation metrics (Fig. 5). The analysis revealed a statistically significant improvement in the primary metric of segmentation overlap, with the average Dice score increasing from 99.506% to **99.519%** ($t = 3.29$, $p = 0.0012$). This enhancement in overall accuracy was underpinned by a pronounced and highly significant increase in segmentation precision, which rose from 99.568% to **99.623%** ($t = 7.00$, $p < 0.0001$), indicating a substantial reduction in false positive voxels. A corresponding trade-off was observed in recall, which saw a small but significant decrease from 99.445% to **99.417%** ($t = 6.26$, $p < 0.0001$), reflecting a slight increase in false negatives.

The impact of DA was particularly evident in the analysis of boundary delineation metrics. While the Hausdorff Distance (HD) and Average Surface Distance (ASD) did not show a statistically significant change, the 95$^{\text{th}}$ percentile Hausdorff Distance (HD95) demonstrated a significant reduction from 0.859 mm to **0.783** mm ($t = 3.10$, $p = 0.0022$). This key finding indicates that the domain adaptation strategy was highly effective at mitigating the most extreme segmentation errors, thereby yielding more consistent and reliable boundary predictions.

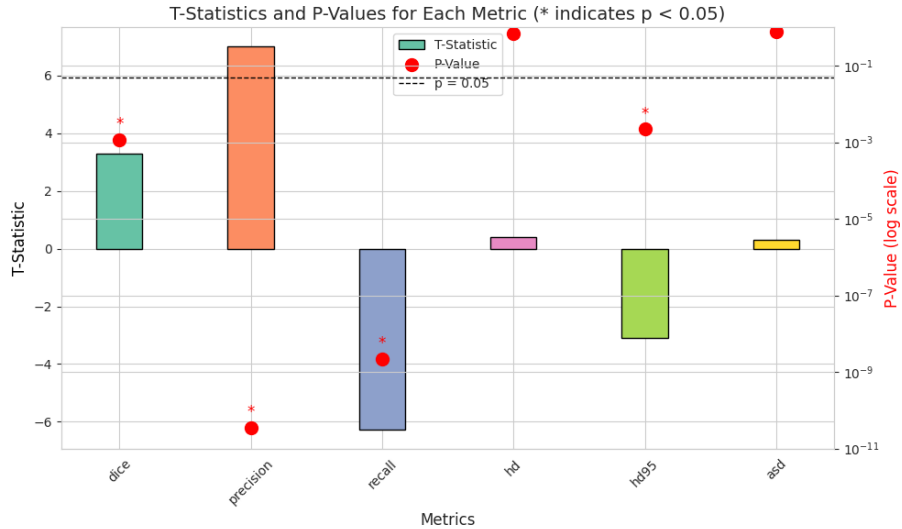

Figure 7: Paired t-statistics and corresponding p-values (log scale) for each segmentation metric between the compared methods. Bars represent the t-statistics for each metric, while red dots indicate the p-values. The dashed horizontal line denotes the significance threshold ($p = 0.05$). Metrics marked with a red asterisk (*) indicate statistically significant differences ($p < 0.05$).

## 5.1 Ablation Study

To systematically evaluate the contribution of different domain adaptation (DA) strategies, we conducted an ablation across four settings: (i) GRL+MMD (Study 1),

(ii) GRL only (Study 2), (iii) MMD only (Study 3), and (iv) no DA (Study 4). For each study, we varied the segmentation loss weighting between Dice cross-entropy (DiceCE) and Focal loss while keeping the DA loss weights fixed where applicable. Validation was assessed using Dice, Precision, Recall, Hausdorff Distance (HD), 95th percentile HD (HD95), and Average Surface Distance (ASD).

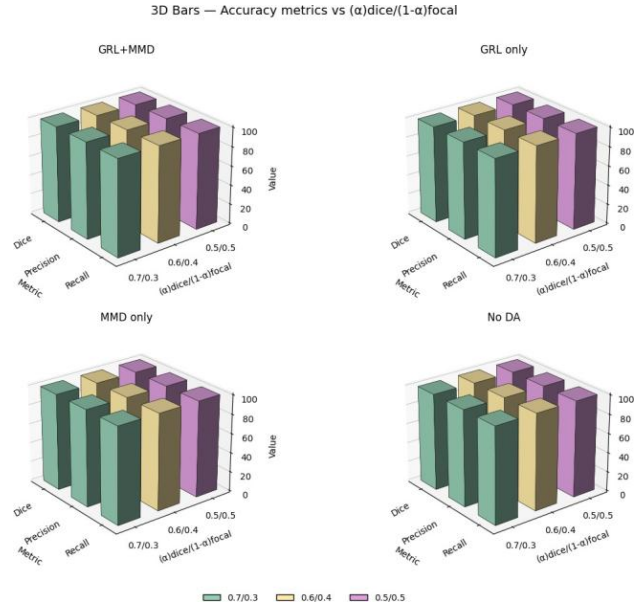

*Figure 8: 3D bar plots of accuracy metrics (Dice, Precision, Recall) vs Dice–Focal loss mixtures across DA settings.*

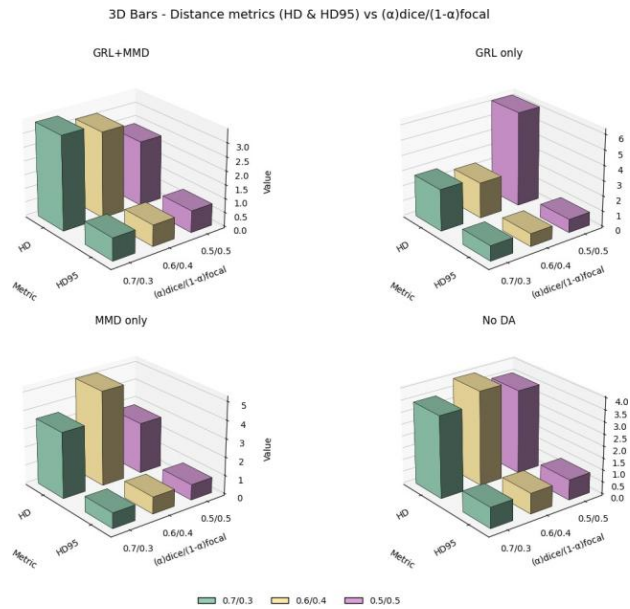

*Figure 9: 3D bar plots of distance metrics (HD, HD95) vs loss mixtures across DA settings.*

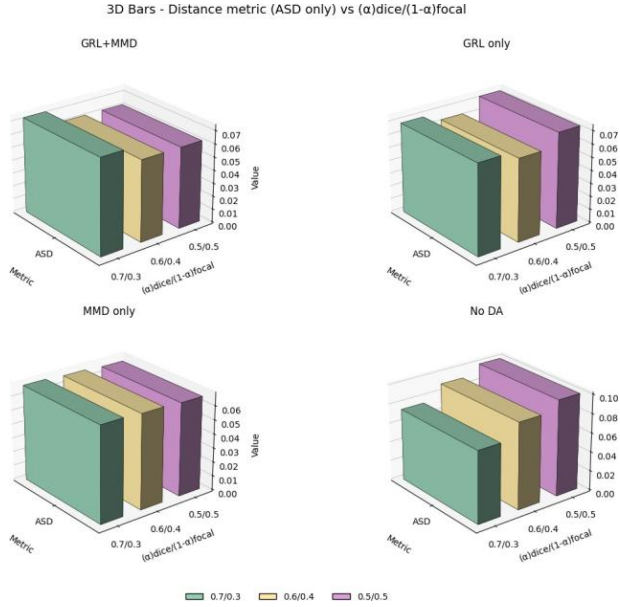

*Figure 10: 3D bar plots of ASD vs loss mixtures across DA settings.*

### 5.1.1 GRL+MMD Combination

As shown in Table 2, combining GRL and MMD consistently improves performance across most metrics. The best configuration was observed with a DiceCE/Focal ratio of 0.6/0.4, which achieved the highest Dice (99.53%), highest Precision (99.64%), and the lowest HD95 (0.77) and ASD (0.062). While a ratio of 0.7/0.3 provided slightly better Recall and HD, the 0.6/0.4 configuration offered the most balanced trade-off, highlighting the complementary role of adversarial alignment and distribution matching.

Table 2: Study 1 (GRL+MMD). Best results are in **bold**.

| DiceCE/Focal | Dice | Precision | Recall | HD ↓ | HD95 ↓ | ASD ↓ |
|---|---|---|---|---|---|---|
| 0.5/0.5 | 99.52 | 99.60 | 99.44 | 3.42 | 0.82 | 0.073 |
| 0.6/0.4 | 99.53 | 99.64 | 99.43 | 3.13 | 0.77 | 0.062 |
| 0.7/0.3 | 99.53 | 99.60 | 99.46 | 2.36 | 0.79 | 0.062 |

### 5.1.2 Single DA Strategies

Using only GRL (Table 3) provided the lowest HD of 2.34 at a 0.6/0.4 ratio, suggesting sharper boundary control. However, Precision and Dice were slightly lower than the GRL+MMD combination. MMD alone (Table 4) yielded modest improvements in HD95 and ASD under the 0.7/0.3 configuration, but overall stability was weaker compared to the combined approach. These results indicate that adversarial alignment (GRL) and distribution matching (MMD) are individually beneficial but not sufficient to consistently optimize all metrics.

Table 3: Study 2 (GRL only).

| DiceCE/Focal | Dice | Precision | Recall | HD ↓ | HD95 ↓ | ASD ↓ |
| --- | --- | --- | --- | --- | --- | --- |
| 0.5/0.5 | 99.49 | 99.51 | 99.47 | 2.74 | 0.99 | 0.069 |
| 0.6/0.4 | 99.52 | 99.63 | 99.42 | 2.34 | 0.84 | 0.063 |
| 0.7/0.3 | 99.51 | 99.61 | 99.42 | 6.22 | 0.87 | 0.073 |

Table 4: Study 3 (MMD only).

| DiceCE/Focal | Dice | Precision | Recall | HD ↓ | HD95 ↓ | ASD ↓ |
| --- | --- | --- | --- | --- | --- | --- |
| 0.5/0.5 | 99.51 | 99.60 | 99.43 | 3.60 | 0.84 | 0.068 |
| 0.6/0.4 | 99.51 | 99.60 | 99.41 | 5.17 | 0.92 | 0.067 |
| 0.7/0.3 | 99.52 | 99.61 | 99.43 | 2.78 | 0.84 | 0.066 |

### 5.1.3 No Domain Adaptation

Finally, the baseline without DA (Table 5) achieved the highest Recall (99.47%) but at the cost of degraded surface accuracy, with larger HD95 (0.87–0.90) and ASD (0.075–0.100). This suggests that while the model becomes more sensitive, it tends to over-segment and produces less precise boundaries. The comparison underscores the importance of DA in maintaining both global accuracy and boundary integrity.

Table 5: Study 4 (No DA).

| DiceCE/Focal | Dice | Precision | Recall | HD ↓ | HD95 ↓ | ASD ↓ |
| --- | --- | --- | --- | --- | --- | --- |
| 0.5/0.5 | 99.50 | 99.53 | 99.47 | 3.45 | 0.87 | 0.075 |
| 0.6/0.4 | 99.51 | 99.57 | 99.45 | 3.97 | 0.86 | 0.090 |
| 0.7/0.3 | 99.51 | 99.58 | 99.44 | 3.53 | 0.85 | 0.100 |

Overall, the combination of GRL and MMD with a DiceCE/Focal ratio of 0.6/0.4 produced the most consistent gains across all validation metrics. GRL alone minimized extreme boundary errors, while MMD alone provided incremental improvements in surface metrics. In contrast, removing DA altogether favored sensitivity (Recall) but compromised structural precision. These results highlight the complementary benefits of adversarial and distribution-based DA methods in achieving robust inter-domain segmentation.

# 6 Discussion

This work demonstrates that transformer-based segmentation models, when equipped with domain adaptation, can achieve both high volumetric accuracy and improved boundary robustness for multi-institutional QCT data. Although baseline performance was already strong, integrating GRL and MMD led to statistically significant improvements in Dice, precision, and HD95, reflecting better contour fidelity and scanner-invariant feature learning. The ablation analysis revealed that GRL sharpened boundary delineation while MMD reduced distributional bias, and their combination provided the most balanced trade-off across overlap and boundary metrics. Importantly, while recall decreased slightly with adaptation, the trade-off favored reduced over-segmentation, suggesting more confident and anatomically precise predictions.

Beyond segmentation, we further validated the consistency of predicted masks by extracting radiomic features and comparing them with ground truth. Correlation analysis across 205 cases revealed near-perfect agreement, with most features showing Pearson's $r > 0.99$. Features such as Busyness, Voxel Volume, Gray Level Non-Uniformity, and Energy exhibited extremely high correlations ($r > 0.9998$), while shape descriptors such as Surface Area and Sphericity showed slightly lower values but remained robust ($r \approx 0.996$ and $r \approx 0.906$, respectively). This indicates that the adapted segmentation not only improves boundary accuracy but also preserves clinically relevant textural and morphological information critical for downstream applications such as finite element structural analysis or fracture risk assessment. Taken together, these findings highlight that PF-DAformer enables population and scanner-invariant segmentation that maintains fidelity of quantitative imaging biomarkers, supporting its potential for reliable deployment in multi-center QCT pipelines.

# 7 Conclusion

The combination of GRL and MMD achieved the most balanced performance, reducing contour deviations and enhancing surface smoothness. GRL improved edge fidelity, while MMD enforced consistent feature alignment, resulting in robust and domain invariant segmentation across scanners.

**Acknowledgement**

This research was in part supported by grants from the National Institutes of Health, USA (U19AG055373, 1R15HL172198, and 1R15HL173852) and American Heart Association (#25AIREA1377168). We gratefully acknowledge Dr. Sundeep Khosla, Dr. Shreyasee Amin, and Elizabeth J. Atkinson at Mayo Clinic, Rochester, MN, for providing the QCT dataset from the Rochester, Minnesota cohort.

**Declaration of Competing Interest**

The authors declare that they have no known competing financial interests or personal relationships that could have appeared to influence the work reported in this paper.